\pdfoutput=1
\documentclass[11pt]{article}

\usepackage[]{naacl2021}

\usepackage{times}
\usepackage{latexsym}

\usepackage[T1]{fontenc}
\usepackage[utf8]{inputenc}

\usepackage{microtype}

\usepackage{booktabs}
\usepackage{amsmath}
\usepackage{multirow}
\usepackage{multicol}
\usepackage{algorithm,algorithmic}
\usepackage{amsfonts}
\usepackage{url}
\usepackage{graphicx}

\title{Comparison of Grammatical Error Correction Using \\Back-Translation Models}

\author{Aomi Koyama ~~ Kengo Hotate\thanks{~~Current affiliation: Recruit Co., Ltd.} ~~ Masahiro Kaneko\thanks{~~Current affiliation: Tokyo Institute of Technology} ~~ Mamoru Komachi \\ Tokyo Metropolitan University \\ \texttt{koyama-aomi@ed.tmu.ac.jp, kengo\_hotate@r.recruit.co.jp} \\ \texttt{masahiro.kaneko@nlp.c.titech.ac.jp, komachi@tmu.ac.jp}}

\begin{document}
\maketitle
\begin{abstract}
Grammatical error correction (GEC) suffers from a lack of sufficient parallel data.
Therefore, GEC studies have developed various methods to generate pseudo data, which comprise pairs of grammatical and artificially produced ungrammatical sentences.
Currently, a mainstream approach to generate pseudo data is back-translation (BT).
Most previous GEC studies using BT have employed the same architecture for both GEC and BT models.
However, GEC models have different correction tendencies depending on their architectures.
Thus, in this study, we compare the correction tendencies of the GEC models trained on pseudo data generated by different BT models, namely, Transformer, CNN, and LSTM.
The results confirm that the correction tendencies for each error type are different for every BT model.
Additionally, we examine the correction tendencies when using a combination of pseudo data generated by different BT models.
As a result, we find that the combination of different BT models improves or interpolates the $\mathrm{F_{0.5}}$ scores of each error type compared with that of single BT models with different seeds.
\end{abstract}

\section{Introduction}
Grammatical error correction (GEC) aims to automatically correct errors in text written by language learners.
It is generally considered as a translation from ungrammatical sentences to grammatical sentences, and GEC studies use machine translation (MT) models as GEC models.
After \citet{yuan-briscoe-2016-grammatical} applied an encoder--decoder (EncDec) model \citep{sutskever-etal-2014-sequence, bahdanau-etal-2015-neural} to GEC, various EncDec-based GEC models have been proposed \citep{ji-etal-2017-nested, chollampatt-ng-2018-multilayer, junczys-dowmunt-etal-2018-approaching, zhao-etal-2019-improving, kaneko-etal-2020-encoder}.

GEC models have different correction tendencies in each architecture.
For example, a GEC model based on CNN \citep{gehring-etal-2017-convolutional} tends to correct errors effectively using the local context \citep{chollampatt-ng-2018-multilayer}.
Furthermore, some studies have combined multiple GEC models to exploit the difference in correction tendencies, thereby improving performance \citep{grundkiewicz-junczys-dowmunt-2018-near, kantor-etal-2019-learning}.

Despite their success, EncDec-based models require considerable amounts of parallel data for training \citep{koehn-knowles-2017-six}.
% However, GEC suffers from a lack of sufficient publicly available parallel data.
However, GEC suffers from a lack of sufficient parallel data.
Accordingly, GEC studies have developed various pseudo data generation methods \citep{xie-etal-2018-noising, ge-etal-2018-fluency, zhao-etal-2019-improving, lichtarge-etal-2019-corpora, xu-etal-2019-erroneous, choe-etal-2019-neural,qiu-etal-2019-improving, grundkiewicz-etal-2019-neural, kiyono-etal-2019-empirical, grundkiewicz-junczys-dowmunt-2019-minimally, wang-etal-2020-controllable, takahashi-etal-2020-grammatical, wang-zheng-2020-improving, zhou-etal-2020-improving-grammatical, wan-etal-2020-improving}.
Moreover, \citet{wan-etal-2020-improving} showed that the correction tendencies of the GEC model are different when using (1) a pseudo data generation method by adding noise to latent representations and (2) a rule-based pseudo data generation method.
Furthermore, they improved the GEC model by combining pseudo data generated by these methods.
Therefore, the combination of pseudo data generated by multiple methods with different tendencies allows us to improve the GEC model further.

One of the most common methods to generate pseudo data is back-translation (BT) \citep{sennrich-etal-2016-improving}.
In BT, we train a BT model (i.e., the reverse model of the GEC model), which generates an ungrammatical sentence from a given grammatical sentence.
Subsequently, a grammatical sentence is provided as an input to the BT model, generating a sentence containing pseudo errors.
Finally, pairs of erroneous sentences and their input sentences are used as pseudo data to train a GEC model.

\citet{kiyono-etal-2019-empirical} reported that a GEC model using BT achieved the best performance among other pseudo data generation methods.
However, most previous GEC studies using BT have used the BT model with the same architecture as the GEC model \citep{xie-etal-2018-noising, ge-etal-2018-fluency, ge-etal-2018-reaching, zhang-etal-2019-sequencetosequence, kiyono-etal-2019-empirical, kiyono-etal-2020-massive}.
Thus, it is unclear whether the correction tendencies differ when using BT models with different architectures.

We investigated correction tendencies of the GEC model using pseudo data generated by different BT models.
Specifically, we used three BT models: Transformer \citep{vaswani-etal-2017-attention}, CNN \citep{gehring-etal-2017-convolutional}, and LSTM \citep{luong-etal-2015-effective}.
The results showed that correction tendencies of each error type are different for each BT model.
In addition, we examined correction tendencies of the GEC model when using a combination of pseudo data generated by different BT models.
As a result, we found that the combination of different BT models improves or interpolates the $\mathrm{F_{0.5}}$ scores of each error type compared with that of single BT models with different seeds.

The main contributions of this study are as follows:
\begin{itemize}
  \item We confirmed that correction tendencies of the GEC model are different for each BT model.
 \item We found that the combination of different BT models improves or interpolates the $\mathrm{F_{0.5}}$ scores compared with that of single BT models with different seeds.
\end{itemize}

\section{Related Works}
\subsection{Back-Translation in Grammatical Error Correction}
\citet{sennrich-etal-2016-improving} showed that BT can effectively improve neural machine translation.
Therefore, many MT studies focused on BT \citep{poncelas-etal-2018-investigating, fadaee-monz-2018-back, edunov-etal-2018-understanding, graca-etal-2019-generalizing, caswell-etal-2019-tagged, edunov-etal-2020-evaluation, soto-etal-2020-selecting, dou-etal-2020-dynamic}.
Subsequently, BT was applied to GEC.
For example, \citet{xie-etal-2018-noising} proposed noising beam search methods, and \citet{ge-etal-2018-fluency} proposed back-boost learning.
Moreover, \citet{rei-etal-2017-artificial} and \citet{kasewa-etal-2018-wronging} applied BT to a grammatical error detection task.

\citet{kiyono-etal-2019-empirical} compared pseudo data generation methods, including BT.
They reported that (1) the GEC model using BT achieved the best performance and (2) using pseudo data for pre-training improves the GEC model more effectively than using a combination of pseudo data and genuine parallel data.
This is because the amount of pseudo data is much larger than that of genuine parallel data.
This usage of pseudo data in GEC contrasts with the usage of a combination of pseudo data and genuine parallel data in MT \citep{sennrich-etal-2016-improving, edunov-etal-2018-understanding, caswell-etal-2019-tagged}.

\citet{htut-tetreault-2019-unbearable} compared four GEC models---Transformer, CNN, PRPN \citep{shen-etal-2018-neural}, and ON-LSTM \citep{shen-etal-2019-ordered}---using pseudo data generated by different BT models.
Specifically, they used Transformer and CNN as BT models.
It was reported that the Transformer using pseudo data generated by CNN achieved the best $\mathrm{F_{0.5}}$ score.
However, the correction tendencies for each BT model were not reported.
Moreover, although using pseudo data for pre-training is common in GEC \citep{zhao-etal-2019-improving, lichtarge-etal-2019-corpora, grundkiewicz-etal-2019-neural, zhou-etal-2020-improving-grammatical, hotate-etal-2020-generating}, they used a less common method of utilizing pseudo data for re-training after training with genuine parallel data.
Therefore, we used Transformer as the GEC model and investigated correction tendencies when using Transformer, CNN, and LSTM as BT models.
Further, we used pseudo data to pre-train the GEC model.

\subsection{Correction Tendencies When Using Each Pseudo Data Generation Method}
\citet{white-rozovskaya-2020-comparative} conducted a comparative study of two rule/probability-based pseudo data generation methods.
The first method \citep{grundkiewicz-etal-2019-neural} generates pseudo data using a confusion set based on a spell checker.
The second method \citep{choe-etal-2019-neural} generates pseudo data using human edits extracted from annotated GEC corpora or replacing prepositions/nouns/verbs with predefined rules.
Based on the comparison results of these methods, it was reported that the former has better performance in correcting spelling errors, whereas the latter has better performance in correcting noun number and tense errors.
In addition, \citet{lichtarge-etal-2019-corpora} compared pseudo data extracted from Wikipedia edit histories with that generated by round-trip translation.
They reported that the former enables better performance in correcting morphology and orthography errors, whereas the latter enables better performance in correcting preposition and pronoun errors.
Similarly, we reported correction tendencies of the GEC model when using pseudo data generated by three BT models with different architectures.

Some studies have used a combination of pseudo data generated by different methods for training the GEC model \citep{lichtarge-etal-2019-corpora, zhou-etal-2020-improving-grammatical, zhou-etal-2020-pseudo, wan-etal-2020-improving}.
For example, \citet{zhou-etal-2020-improving-grammatical} proposed a pseudo data generation method that pairs sentences translated by statistical machine translation and neural machine translation.
Then, they combined pseudo data generated by it with pseudo data generated by BT to pre-train the GEC model.
However, they did not report the correction tendencies of the GEC model when using combined pseudo data.
Conversely, we reported correction tendencies when using a combination of pseudo data generated by different BT models.

\begin{table}[t]
\center
\small
\begin{tabular}{lrcc}
      \toprule
      Dataset&Sents.&Refs.&Split\\
      \toprule
      BEA-train&564,684&1&train\\
      BEA-valid&4,384&1&valid\\
      \midrule
      CoNLL-2014&1,312&2&test\\
      JFLEG&747&4&test\\
      BEA-test&4,477&5&test\\
      \midrule
      Wikipedia&9,000,000&-&-\\
      \bottomrule
\end{tabular}
\caption{Dataset used in the experiments.}
\label{tb:dataset_summary}
\end{table}

\begin{table*}[t]
\center
\scalebox{0.78}{
\begin{tabular}{lrccccccc}
      \toprule
      &&\multicolumn{3}{c}{CoNLL-2014}&JFLEG&\multicolumn{3}{c}{BEA-test}\\
      \cmidrule(lr){3-5}\cmidrule(lr){6-6}\cmidrule(lr){7-9}
      Back-translation model & Pseudo data & Prec. & Rec. & $\mathrm{F_{0.5}}$ & GLEU & Prec. & Rec. & $\mathrm{F_{0.5}}$\\
      \toprule
      None (Baseline) & - & 58.5/65.8 & 31.3/31.5 & 49.8/54.0 & 53.0/53.7 & 52.6/61.4 & 42.8/42.8 & 50.2/56.5\\
      Transformer & 9M &\textbf{65.0}/68.6 & \textbf{37.6}/\textbf{37.7} & \textbf{56.7}/\textbf{59.0} & 57.7/58.3 & 61.1/66.5 & 49.8/50.7 & 58.4/62.6\\
      CNN & 9M & 64.0/68.1 & 37.4/37.4 & 56.0/58.5 & \textbf{57.8}/\textbf{58.4} & \textbf{61.9}/\textbf{67.5} & \textbf{50.7}/\textbf{51.0} & \textbf{59.3}/\textbf{63.4}\\
      LSTM & 9M & 64.7/\textbf{68.8} & 36.2/36.4 & 55.9/58.4 & 57.0/57.4 & 61.3/67.1 & 49.5/49.9 & 58.5/62.8\\
      \midrule
      Transformer \& CNN & 18M & 65.2/\textbf{69.1} & \textbf{38.7}/\textbf{39.1} & \textbf{57.3}/\textbf{59.9} & \textbf{57.9}/58.5 & \textbf{63.1}/\textbf{67.6} & 51.1/51.1 & \textbf{60.2}/\textbf{63.5}\\
      Transformer \& Transformer & 18M & 65.5/68.3 & 37.9/38.0 & 57.2/58.9 & 57.5/58.0 & 63.0/67.0 & 51.0/50.7 & \textbf{60.2}/63.0\\
      CNN \& CNN & 18M & \textbf{65.6}/\textbf{69.1} & 38.2/38.7 & \textbf{57.3}/59.8 & \textbf{57.9}/\textbf{58.6} & 61.9/67.1 & \textbf{51.4}/\textbf{51.6} & 59.5/63.3\\
      \bottomrule
\end{tabular}
}
\caption{Results of each GEC model. The left and right scores represent single and ensemble model results, respectively. The top group delineates the performance of the GEC model using each BT model, and the bottom group delineates the performance of the GEC model when using combined pseudo data.}
\label{tb:score}
\end{table*}

\section{Experimental Setup}
\subsection{Dataset}
Table \ref{tb:dataset_summary} shows the details of the dataset used in the experiments.
We used the BEA-2019 workshop official shared task dataset \citep{bryant-etal-2019-bea} as the training and validation data.
This dataset consists of FCE \citep{yannakoudakis-etal-2011-new}, Lang-8 \citep{mizumoto-etal-2011-mining, tajiri-etal-2012-tense}, NUCLE \citep{dahlmeier-etal-2013-building}, and W\&I+LOCNESS \citep{granger-1998-computerized, yannakoudakis-etal-2018-developing}.
Following \citet{chollampatt-ng-2018-multilayer}, we removed sentence pairs with identical source and target sentences from the training data.
Next, we applied byte pair encoding \cite{sennrich-etal-2016-neural} to both source and target sentences.
Here, we acquired subwords from the target sentences in the training data and set the vocabulary size to 8,000.
Hereinafter, we refer to the training and validation data as BEA-train and BEA-valid, respectively.

% We used Wikipedia\footnote{We used the 2020-07-06 dump file at \url{https://dumps.wikimedia.org/other/cirrussearch/}.} as a seed corpus to generate pseudo data and removed sentences that were possibly inappropriate, such as URLs.
We used Wikipedia\footnote{We used the 2020-07-06 dump file at \url{https://dumps.wikimedia.org/other/cirrussearch/}.} as a seed corpus to generate pseudo data and removed possibly inappropriate sentences, such as URLs.
In total, we extracted 9M sentences randomly.

\subsection{Evaluation}
We evaluated the CoNLL-2014 test set (CoNLL-2014) \citep{ng-etal-2014-conll}, the JFLEG test set (JFLEG) \citep{heilman-etal-2014-predicting, napoles-etal-2017-jfleg}, and the official test set of the BEA-2019 shared task (BEA-test).
We reported $\mathrm{M^2}$ \citep{dahlmeier-ng-2012-better} for the CoNLL-2014 and GLEU \citep{napoles-etal-2015-ground, napoles-etal-2016-gleu} for the JFLEG.
We also reported the scores measured by ERRANT \citep{felice-etal-2016-automatic, bryant-etal-2017-automatic} for the BEA-valid and BEA-test.
All the reported results, except for the ensemble model, are the average of three distinct trials using three different random seeds\footnote{To reduce the influence of the BT model's seed, we prepared BT models trained with the corresponding seed of each GEC model. Then, we pre-trained each GEC model using pseudo data generated by the corresponding BT models.}.
In the ensemble model, we reported the ensemble results of the three GEC models.

\subsection{Grammatical Error Correction Model}
Following \citet{kiyono-etal-2019-empirical}, we adopted Transformer, which is a representative EncDec-based model, using the fairseq toolkit \citep{ott-etal-2019-fairseq}.
We used the ``Transformer (base)'' settings of \citet{vaswani-etal-2017-attention}\footnote{Considering the limitation of computing resources, we used ``Transformer (base)'' instead of ``Transformer (big)''.}, which has a 6-layer encoder and decoder with a dimensionality of 512 for both input and output and 2,048 for inner-layers, and 8 self-attention heads.
We pre-trained GEC models on each 9M pseudo data generated by each BT model\footnote{See Section \ref{bt_model} for details of the BT models.} and then fine-tuned them on BEA-train.
We optimized the model by using Adam \citep{kingma-ba-2015-adam} in pre-training and with Adafactor \citep{shazeer-stern-2018-adafactor} in fine-tuning.
Most of the hyperparameter settings were the same as those described in \citet{kiyono-etal-2019-empirical}.
Additionally, we trained a GEC model using only the BEA-train without pre-training as a baseline model.

We investigated correction tendencies when using a combination of pseudo data generated by different BT models.
Therefore, we pre-trained a GEC model on combined pseudo data and then fine-tuned it on the BEA-train.
Notably, in this experiment, we combined pseudo data generated by the Transformer and CNN because they improved the GEC models compared with LSTM in most cases (Section \ref{overall_results}).
Specifically, we obtained 9M pseudo data from the Transformer and CNN and then created 18M pseudo data by combining them.
To eliminate the effect of increasing the pseudo data amount, we prepared GEC models that used a combination of pseudo data generated by single BT models with different seeds.
We provided all BT models with the same target sentences to focus on the difference in the pseudo source sentences.
Hence, in the combined pseudo data, the number of source sentence types increases; however, the number of target sentence types does not increase.

\subsection{Back-Translation Model}
\label{bt_model}
Based on the GEC studies that used BT, we selected the Transformer \citep{vaswani-etal-2017-attention}, CNN \citep{gehring-etal-2017-convolutional}, and LSTM \citep{luong-etal-2015-effective}.
For all BT models, we used implementations of the fairseq toolkit and its default settings, except for common settings\footnote{When training each BT model, the argument \textit{--arch} in the fairseq toolkit was set to \texttt{transformer}, \texttt{fconv}, and \texttt{lstm} for the Transformer, CNN, and LSTM, respectively.}.
\paragraph{Common settings.}
We used the Adam optimizer with $\beta_{\mathrm{1}} = 0.9$ and $\beta_{\mathrm{2}} = 0.98$.
We used label smoothed cross-entropy \citep{szegedy-etal-2016-rethinking} as a loss function and selected the model that achieved the smallest loss on the BEA-valid.
We set the maximum number of epochs to 40.
The learning rate schedule is the same as that described in \citet{vaswani-etal-2017-attention}.
We applied dropout \cite{srivastava-etal-2014-dropout} with a rate of 0.3.
We set the beam size to 5 with length normalization.
Moreover, to generate various errors, we used the noising beam search method proposed by \citet{xie-etal-2018-noising}.
In this method, we add $r\beta_{\mathrm{random}}$ to the score of each hypothesis in the beam search.
Here, $r$ is randomly sampled from a uniform distribution of interval $[0, 1]$, and $\beta_{\mathrm{random}} \in\mathbb{R}_{\ge0}$ is a hyperparameter that adjusts the noise scale.
In this experiment, $\beta_{\mathrm{random}}$ was set to 8, 10, and 12 for the Transformer, CNN, and LSTM, respectively\footnote{Each $\beta_{\mathrm{random}}$ achieved the best $\mathrm{F_{0.5}}$ score on the BEA-valid in the preliminary experiments.}.
\paragraph{Transformer.}
Our Transformer model was based on \citet{vaswani-etal-2017-attention}, which is a 6-layer encoder and decoder with 512-dimensional embeddings, 2,048 for inner-layers, and 8 self-attention heads.
\paragraph{CNN.}
Our CNN model was based on \citet{gehring-etal-2017-convolutional}, which is a 20-layer encoder and decoder with 512-dimensional embeddings, both using kernels of width 3 and hidden size 512.
\paragraph{LSTM.}
Our LSTM model was based on \citet{luong-etal-2015-effective}, which is a 1-layer encoder and decoder with 512-dimensional embeddings and hidden size 512.

\begin{table*}[t]
\center
\scalebox{0.78}{
\begin{tabular}{lccccc|ccc}
      \toprule
      &&&\multicolumn{6}{c}{Back-translation model}\\
      \cmidrule(lr){4-9}
      \multirow{2}{*}{Error type} & \multirow{2}{*}{Freq.} & \multirow{2}{*}{Baseline} &  \multirow{2}{*}{Transformer} & \multirow{2}{*}{CNN} & \multirow{2}{*}{LSTM} & \multicolumn{1}{l}{Transformer} & \multicolumn{1}{l}{Transformer} & \multicolumn{1}{l}{CNN}\\
      &&&&&&\multicolumn{1}{l}{\& CNN}&\multicolumn{1}{l}{\& Transformer}& \multicolumn{1}{l}{\& CNN} \\
      \toprule
      OTHER & 697 & 22.2$\pm$1.77 & \textbf{31.8$\pm$0.71} & 31.7$\pm$0.77 & 30.6$\pm$0.16 & \textbf{34.2$\pm$1.03} & 31.8$\pm$1.01 & 31.6$\pm$0.74\\
      PUNCT & 613 & 65.6$\pm$2.02 & 64.6$\pm$0.42 & \textbf{67.8$\pm$0.83} & 67.3$\pm$1.83 & 65.9$\pm$1.51 & 66.0$\pm$0.73 & \textbf{67.8$\pm$0.93}\\
      DET & 607 & 53.8$\pm$0.71 & 64.8$\pm$1.62 & 65.0$\pm$0.41 & \textbf{65.2$\pm$0.83} & 64.8$\pm$0.64 & \textbf{66.7$\pm$1.15} & 64.7$\pm$0.75\\
      PREP & 417 & 48.2$\pm$0.55 & 58.1$\pm$0.76 & \textbf{59.3$\pm$0.54} & 55.2$\pm$1.74 & \textbf{61.1$\pm$0.43} & 60.3$\pm$0.76 & 60.3$\pm$1.06\\
      ORTH & 381 & 72.7$\pm$2.47 & 77.2$\pm$0.50 & \textbf{78.7$\pm$1.50} & 78.0$\pm$1.95 & \textbf{79.2$\pm$1.25} & 78.4$\pm$1.28 & 78.8$\pm$0.74\\
      SPELL & 315 & 58.3$\pm$3.49 & 71.0$\pm$1.71 & 71.1$\pm$1.45 & \textbf{71.6$\pm$0.50} & \textbf{73.3$\pm$1.03} & 72.5$\pm$0.40 & 71.1$\pm$0.49\\
      NOUN:NUM & 263 & 57.8$\pm$2.23 & \textbf{64.4$\pm$1.09} & 63.7$\pm$0.90 & 63.9$\pm$1.35 & 66.2$\pm$0.43 & \textbf{66.3$\pm$0.61} & 64.6$\pm$1.41\\
      VERB:TENSE & 256 & 43.9$\pm$2.35 & 52.1$\pm$1.58 & \textbf{54.6$\pm$0.94} & 52.6$\pm$0.50 & 53.7$\pm$1.71 & 54.6$\pm$0.64 & \textbf{54.8$\pm$1.27}\\
      VERB:FORM & 213 & 62.0$\pm$2.26 & 66.7$\pm$2.63 & \textbf{67.1$\pm$0.46} & 66.0$\pm$1.60 & 66.3$\pm$0.34 & \textbf{66.9$\pm$1.54} & 66.6$\pm$1.01\\
      VERB & 196 & 32.5$\pm$3.41 & 36.0$\pm$1.18 & 36.3$\pm$0.91 & \textbf{39.7$\pm$3.05} & \textbf{42.7$\pm$3.83} & 39.0$\pm$0.76 & 38.2$\pm$0.98\\
      VERB:SVA & 157 & 66.1$\pm$1.38 & 73.7$\pm$3.00 & \textbf{75.6$\pm$0.86} & 73.8$\pm$2.51 & 75.1$\pm$1.04 & \textbf{76.3$\pm$1.20} & 74.3$\pm$0.44\\
      MORPH & 155 & 54.0$\pm$2.03 & 61.9$\pm$1.97 & \textbf{63.8$\pm$1.23} & \textbf{63.8$\pm$0.53} & 64.5$\pm$0.62 & \textbf{66.3$\pm$1.26} & 63.8$\pm$2.84\\
      PRON & 139 & 43.8$\pm$2.00 & \textbf{53.0$\pm$2.79} & 51.8$\pm$0.14 & 49.6$\pm$1.93 & \textbf{53.3$\pm$1.10} & 52.7$\pm$2.75 & \textbf{53.3$\pm$0.46}\\
      NOUN & 129 & 19.7$\pm$2.04 & \textbf{31.4$\pm$0.62} & 30.2$\pm$2.39 & 30.5$\pm$2.17 & \textbf{35.9$\pm$2.90} & 34.5$\pm$1.48 & 32.8$\pm$2.80\\
      \bottomrule
\end{tabular}
}
\caption{Each error type's $\mathrm{F_{0.5}}$ of the single models on the BEA-test. We extracted error types with a frequency of 100 or more. The total frequency of all error types was 4,882. For details of error types, see \citet{bryant-etal-2017-automatic}. }
\label{tb:error_type_score}
\end{table*}

\section{Results}
\subsection{Overall Results}
\label{overall_results}
\paragraph{Separate pseudo data.}
The top group in Table \ref{tb:score} depicts the results of the GEC model using each BT model; the best BT model was different for each test set.
% Specifically, the GEC model using the Transformer achieved the best scores in the CoNLL-2014.
The GEC model using the Transformer achieved the best scores in the CoNLL-2014.
In contrast, in the JFLEG and BEA-test, the GEC model using CNN achieved the best scores.
Moreover, the GEC model using LSTM achieved a higher $\mathrm{F_{0.5}}$ than that using the Transformer in the BEA-test.
These results suggest that the Transformer, which is robust as the GEC model \citep{kiyono-etal-2019-empirical}, is not necessarily a good BT model.

\paragraph{Combined pseudo data.}
The bottom group of Table \ref{tb:score} shows the results of the GEC model using combined pseudo data.
As shown in Table \ref{tb:score}, a combination of pseudo data generated by different BT models consistently improved the performance compared with pseudo data from a single source (Transformer \& CNN > Transformer, CNN).
In contrast, in some of the items in Table \ref{tb:score}, the performances of the GEC models using the single BT models with different seeds were lower than that using only a single BT model.
For example, when using the Transformer as the BT model, the $\mathrm{F_{0.5}}$ score of the ensemble model using a single BT model was 59.0 on the CoNLL-2014, whereas that using two homogeneous BT models was 58.9 (Transformer \& Transformer: 58.9 < Transformer: 59.0).
Similarly, for CNN, the $\mathrm{F_{0.5}}$ score of the ensemble model using only a single BT model was 63.4 on the BEA-test, whereas that using two homogeneous BT models was 63.3 (CNN \& CNN: 63.3 < CNN: 63.4).
Hence, the combination of different BT models enables the construction of a more robust GEC model than the combination of single BT models with different seeds.

\subsection{Results of Each Error Type}
\paragraph{Separate pseudo data.}
The left side of Table \ref{tb:error_type_score} illustrates the $\mathrm{F_{0.5}}$ scores of the single models on the BEA-test across various error types.
% When using the Transformer as the BT model, the performance of PRON (pronoun) was high.
% In contrast, the performance of PREP (preposition), VERB:TENSE, and VERB:SVA (subject-verb agreement) was high when using CNN, and the performance of VERB was high when using LSTM, to name a few.
When using the Transformer as the BT model, the performance of PRON was high.
In contrast, the performance of PREP, VERB:TENSE, and VERB:SVA was high when using CNN, and the performance of VERB was high when using LSTM, to name a few.
Therefore, it is considered that correction tendencies of each error type are different depending on the BT model. 

% In PUNCT (punctuation), the performance of the GEC model using the Transformer was lower than that of the baseline model (Transformer: 64.6 < Baseline: 65.6).
In PUNCT, the performance of the GEC model using the Transformer was lower than that of the baseline model (Transformer: 64.6 < Baseline: 65.6).
Moreover, when using CNN and LSTM as the BT model, the performance of PUNCT improved by only approximately 2 points in $\mathrm{F_{0.5}}$ from the baseline model (CNN: 67.8, LSTM: 67.3 > Baseline: 65.6).
It can be seen that this improvement of PUNCT is small compared with that of other error types.
Therefore, when using pseudo data generated by BT, PUNCT is considered an error type that is difficult to improve.

\begin{table*}[t]
\center
\scalebox{0.65}{
\begin{tabular}{lrrrrrrrrrrrr}
      \toprule
      & \multicolumn{4}{c}{Transformer} & \multicolumn{4}{c}{CNN} & \multicolumn{4}{c}{LSTM}\\
      \cmidrule(lr){2-5}\cmidrule(lr){6-9}\cmidrule(lr){10-13}
      \multirow{2}{*}{Error type}& \multicolumn{1}{c}{\multirow{2}{*}{Token}} & \multicolumn{1}{c}{\multirow{2}{*}{Type}} & \multicolumn{1}{l}{$\mathrm{F_{0.5}}$} & \multicolumn{1}{l}{$\mathrm{F_{0.5}}$} & \multicolumn{1}{c}{\multirow{2}{*}{Token}} & \multicolumn{1}{c}{\multirow{2}{*}{Type}} & \multicolumn{1}{l}{$\mathrm{F_{0.5}}$} & \multicolumn{1}{l}{$\mathrm{F_{0.5}}$} & \multicolumn{1}{c}{\multirow{2}{*}{Token}} & \multicolumn{1}{c}{\multirow{2}{*}{Type}} & \multicolumn{1}{l}{$\mathrm{F_{0.5}}$} & \multicolumn{1}{l}{$\mathrm{F_{0.5}}$}\\
      &&&\multicolumn{1}{l}{w/ FT}&\multicolumn{1}{l}{w/o FT}&&&\multicolumn{1}{l}{w/ FT}&\multicolumn{1}{l}{w/o FT}&&&\multicolumn{1}{l}{w/ FT}&\multicolumn{1}{l}{w/o FT}\\
      \toprule
      Overall & 64,733,183 & 12,364,575 & 58.4 & \textbf{32.7} & 77,784,638 & 17,711,223 & \textbf{59.3} & 31.4 & \textbf{90,205,852} & \textbf{25,502,133} & 58.5 & 25.2\\
      \midrule
      OTHER & 16,463,382 & 6,084,184 & \textbf{31.8} & \textbf{10.0} & 20,237,776 & 8,453,119 & 31.7 & 9.6 & \textbf{29,286,403} & \textbf{13,844,773} & 30.6 & 6.3\\
      PUNCT & 3,716,117 & 37,360 & 64.6 & \textbf{47.1} & 3,814,449 & 46,724 & \textbf{67.8} & 46.1 & \textbf{4,082,594} & \textbf{53,739} & 67.3 & 43.1\\
      DET & 8,074,615 & \textbf{39,606} & 64.8 & \textbf{41.5} & \textbf{8,491,264} & 39,402 & 65.0 & 39.2 & 8,217,106 & 33,389 & \textbf{65.2} & 33.0\\
      PREP & 6,832,627 & 19,521 & 58.1 & \textbf{36.7} & 7,935,894 & 23,564 & \textbf{59.3} & 35.8 & \textbf{8,091,043} & \textbf{25,923} & 55.2 & 30.3\\
      ORTH & 3,378,022 & 521,032 & 77.2 & \textbf{62.7} & \textbf{3,973,439} & \textbf{646,475} & \textbf{78.7} & 61.4 & 3,587,805 & 513,787 & 78.0 & 60.0\\
      SPELL & 6,620,395 & 2,795,425 & 71.0 & \textbf{57.0} & 11,224,522 & 4,737,493 & 71.1 & 56.3 & \textbf{11,342,223} & \textbf{5,643,091} & \textbf{71.6} & 50.8\\
      NOUN:NUM & \textbf{2,241,413} & \textbf{31,939} & \textbf{64.4} & \textbf{45.2} & 2,149,748 & 30,205 & 63.7 & 43.9 & 2,177,546 & 28,226 & 63.9 & 41.3\\
      VERB:TENSE & 2,585,017 & 58,935 & 52.1 & \textbf{27.2} & \textbf{2,599,663} & \textbf{60,266} & \textbf{54.6} & 26.6 & 2,418,040 & 59,207 & 52.6 & 22.6\\
      VERB:FORM & 1,287,912 & 47,071 & 66.7 & 45.5 & 1,421,381 & \textbf{48,776} & \textbf{67.1} & \textbf{46.2} & \textbf{1,517,365} & 48,117 & 66.0 & 41.1\\
      VERB & 1,821,117 & 328,147 & 36.0 & \textbf{18.5} & 2,201,360 & 453,181 & 36.3 & 17.2 & \textbf{2,704,117} & \textbf{647,785} & \textbf{39.7} & 12.9\\
      VERB:SVA & 761,768 & \textbf{6,564} & 73.7 & 52.5 & 784,762 & 6,136 & \textbf{75.6} & \textbf{52.8} & \textbf{824,241} & 6,019 & 73.8 & 45.5\\
      MORPH & 2,306,204 & 148,506 & 61.9 & 32.5 & 2,308,793 & 147,657 & \textbf{63.8} & \textbf{32.6} & \textbf{2,613,870} & \textbf{167,440} & \textbf{63.8} & 29.2\\
      PRON & 810,875 & 3,642 & \textbf{53.0} & \textbf{14.7} & 995,686 & 4,013 & 51.8 & 12.7 & \textbf{1,248,554} & \textbf{5,267} & 49.6 & 10.9\\
      NOUN & 4,402,909 & 1,888,994 & \textbf{31.4} & \textbf{14.8} & 6,155,680 & 2,697,991 & 30.2 & 14.4 & \textbf{8,196,758} & \textbf{4,032,482} & 30.5 & 9.8\\
      \bottomrule
\end{tabular}
}
\caption{Number of edit pair tokens and types in pseudo data generated by each BT model and each error type's $\mathrm{F_{0.5}}$ of the single models with and without fine-tuning on the BEA-test. As with Table \ref{tb:error_type_score}, we extracted error types with a frequency of 100 or more in the BEA-test. FT denotes fine-tuning.}
\label{tb:pseudo_data}
\end{table*}

\paragraph{Combined pseudo data.}
The right side of Table \ref{tb:error_type_score} shows the $\mathrm{F_{0.5}}$ scores of the single models using combined pseudo data on the BEA-test across various error types.
Except for 3 of the 14 error types shown in Table \ref{tb:error_type_score}, the GEC model using Transformer \& CNN yielded the higher $\mathrm{F_{0.5}}$ scores than using at least either Transformer \& Transformer or CNN \& CNN.
Therefore, it is considered that the combination of different BT models improves or interpolates performance compared with that of single BT models with different seeds.

In OTHER, the combination of single BT models with different seeds did not improve the performance of OTHER compared with a single BT model (Transformer \& Transformer: 31.8 = Transformer: 31.8 and CNN \& CNN: 31.6 < CNN: 31.7).
Conversely, the combination of different BT models improved the performance of OTHER compared with a single BT model (Transformer \& CNN: 34.2 > Transformer: 31.8, CNN: 31.7).
Thus, by using different BT models, the GEC model is expected to correct more diverse error types.

\paragraph{Effects of different seeds.}
Here, we consider the effect of different seeds in the BT model.
In some error types in Table \ref{tb:error_type_score}, the GEC model using single BT models with different seeds has the higher $\mathrm{F_{0.5}}$ score than that using different BT models.
One of the reasons for this is that there exists some variation (i.e., high standard deviation) in the $\mathrm{F_{0.5}}$ score of each error type, even when changing merely the seed of the BT model.
% For example, in the GEC model using the Transformer, the standard deviation of DET (determiner) was 1.62, which is relatively high.
For example, in the GEC model using the Transformer, the standard deviation of DET was 1.62, which is relatively high.
Then, the $\mathrm{F_{0.5}}$ score of DET using Transformer \& Transformer was higher than that using Transformer \& CNN.
Thus, in error types with some variation, using single BT models with different seeds may improve performance compared with using different BT models.

\section{Discussion}
We examined the number of edit pairs in pseudo data generated by each BT model.
We annotated pseudo data using ERRANT and extracted edit pairs from the pseudo source sentences and target sentences.
Table \ref{tb:pseudo_data} shows the number of edit pair tokens and types in the pseudo data generated by each BT model.
We expected that the higher the number of errors in each error type, the better the $\mathrm{F_{0.5}}$ score of the GEC model for each error type.
However, the results did not show such a tendency.
Specifically, when the number of edit pair tokens and types was the highest in each error type, only 6 of the 14 error types had the highest $\mathrm{F_{0.5}}$ score (ORTH, SPELL, NOUN:NUM, VERB:TENSE, VERB, and MORPH).
This fact implies that simply increasing the number of tokens or types in each error type may not improve each error type's performance in the GEC model.

Moreover, we investigated the performance of the GEC model with and without fine-tuning.
As shown in Table \ref{tb:pseudo_data}, when fine-tuning was not carried out (i.e., pre-training only), the GEC model using the Transformer had the highest $\mathrm{F_{0.5}}$ score, and there was a 7.5 point difference in $\mathrm{F_{0.5}}$ between the Transformer and the LSTM (Transformer: 32.7 > LSTM: 25.2).
However, interestingly, when fine-tuning was performed, the GEC model using LSTM achieved a better $\mathrm{F_{0.5}}$ score than that using the Transformer (Transformer: 58.4 < LSTM: 58.5).
This result suggests that even if the performance of the GEC model is low in pre-training, it may become high after fine-tuning.

\section{Conclusions}
In this study, we investigated correction tendencies based on each BT model.
The results showed that the correction tendencies of each error type varied depending on the BT models.
In addition, we found that the combination of different BT models improves or interpolates the $\mathrm{F_{0.5}}$ score compared with that of single BT models with different seeds.

\section*{Acknowledgments}
We would like to thank Lang-8, Inc. for providing the text data.
We would also like to thank the anonymous reviewers for their valuable comments.
This work was partly supported by JSPS KAKENHI Grant Number 19KK0286.

\bibliography{main_custom}

\begin{thebibliography}{69}
\expandafter\ifx\csname natexlab\endcsname\relax\def\natexlab#1{#1}\fi

\bibitem[{Bahdanau et~al.(2015)Bahdanau, Cho, and
  Bengio}]{bahdanau-etal-2015-neural}
Dzmitry Bahdanau, Kyunghyun Cho, and Yoshua Bengio. 2015.
\newblock \href {http://arxiv.org/abs/1409.0473} {{Neural Machine Translation
  by Jointly Learning to Align and Translate}}.
\newblock In \emph{Proceedings of the 3rd International Conference on Learning
  Representations}, San Diego, California.

\bibitem[{Bryant et~al.(2019)Bryant, Felice, Andersen, and
  Briscoe}]{bryant-etal-2019-bea}
Christopher Bryant, Mariano Felice, {\O}istein~E. Andersen, and Ted Briscoe.
  2019.
\newblock \href {https://doi.org/10.18653/v1/W19-4406} {{The {BEA}-2019 Shared
  Task on Grammatical Error Correction}}.
\newblock In \emph{Proceedings of the Fourteenth Workshop on Innovative Use of
  NLP for Building Educational Applications}, pages 52--75, Florence, Italy.
  Association for Computational Linguistics.

\bibitem[{Bryant et~al.(2017)Bryant, Felice, and
  Briscoe}]{bryant-etal-2017-automatic}
Christopher Bryant, Mariano Felice, and Ted Briscoe. 2017.
\newblock \href {https://doi.org/10.18653/v1/P17-1074} {{Automatic Annotation
  and Evaluation of Error Types for Grammatical Error Correction}}.
\newblock In \emph{Proceedings of the 55th Annual Meeting of the Association
  for Computational Linguistics}, pages 793--805, Vancouver, Canada.
  Association for Computational Linguistics.

\bibitem[{Caswell et~al.(2019)Caswell, Chelba, and
  Grangier}]{caswell-etal-2019-tagged}
Isaac Caswell, Ciprian Chelba, and David Grangier. 2019.
\newblock \href {https://doi.org/10.18653/v1/W19-5206} {{Tagged
  Back-Translation}}.
\newblock In \emph{Proceedings of the Fourth Conference on Machine
  Translation}, pages 53--63, Florence, Italy. Association for Computational
  Linguistics.

\bibitem[{Choe et~al.(2019)Choe, Ham, Park, and Yoon}]{choe-etal-2019-neural}
Yo~Joong Choe, Jiyeon Ham, Kyubyong Park, and Yeoil Yoon. 2019.
\newblock \href {https://doi.org/10.18653/v1/W19-4423} {{A Neural Grammatical
  Error Correction System Built On Better Pre-training and Sequential Transfer
  Learning}}.
\newblock In \emph{Proceedings of the Fourteenth Workshop on Innovative Use of
  NLP for Building Educational Applications}, pages 213--227, Florence, Italy.
  Association for Computational Linguistics.

\bibitem[{Chollampatt and Ng(2018)}]{chollampatt-ng-2018-multilayer}
Shamil Chollampatt and Hwee~Tou Ng. 2018.
\newblock \href
  {https://www.aaai.org/ocs/index.php/AAAI/AAAI18/paper/view/17308} {{A
  Multilayer Convolutional Encoder-Decoder Neural Network for Grammatical Error
  Correction}}.
\newblock In \emph{Proceedings of the Thirty-Second AAAI Conference on
  Artificial Intelligence}, pages 5755--5762, New Orleans, Louisiana.
  Association for the Advancement of Artificial Intelligence.

\bibitem[{Dahlmeier and Ng(2012)}]{dahlmeier-ng-2012-better}
Daniel Dahlmeier and Hwee~Tou Ng. 2012.
\newblock \href {https://www.aclweb.org/anthology/N12-1067} {{Better Evaluation
  for Grammatical Error Correction}}.
\newblock In \emph{Proceedings of the 2012 Conference of the North {A}merican
  Chapter of the Association for Computational Linguistics: Human Language
  Technologies}, pages 568--572, Montr{\'e}al, Canada. Association for
  Computational Linguistics.

\bibitem[{Dahlmeier et~al.(2013)Dahlmeier, Ng, and
  Wu}]{dahlmeier-etal-2013-building}
Daniel Dahlmeier, Hwee~Tou Ng, and Siew~Mei Wu. 2013.
\newblock \href {https://www.aclweb.org/anthology/W13-1703} {{Building a Large
  Annotated Corpus of Learner {E}nglish: The {NUS} Corpus of Learner
  {E}nglish}}.
\newblock In \emph{Proceedings of the Eighth Workshop on Innovative Use of
  {NLP} for Building Educational Applications}, pages 22--31, Atlanta, Georgia.
  Association for Computational Linguistics.

\bibitem[{Dou et~al.(2020)Dou, Anastasopoulos, and
  Neubig}]{dou-etal-2020-dynamic}
Zi-Yi Dou, Antonios Anastasopoulos, and Graham Neubig. 2020.
\newblock \href {https://doi.org/10.18653/v1/2020.emnlp-main.475} {{Dynamic
  Data Selection and Weighting for Iterative Back-Translation}}.
\newblock In \emph{Proceedings of the 2020 Conference on Empirical Methods in
  Natural Language Processing}, pages 5894--5904, Online. Association for
  Computational Linguistics.

\bibitem[{Edunov et~al.(2018)Edunov, Ott, Auli, and
  Grangier}]{edunov-etal-2018-understanding}
Sergey Edunov, Myle Ott, Michael Auli, and David Grangier. 2018.
\newblock \href {https://doi.org/10.18653/v1/D18-1045} {{Understanding
  Back-Translation at Scale}}.
\newblock In \emph{Proceedings of the 2018 Conference on Empirical Methods in
  Natural Language Processing}, pages 489--500, Brussels, Belgium. Association
  for Computational Linguistics.

\bibitem[{Edunov et~al.(2020)Edunov, Ott, Ranzato, and
  Auli}]{edunov-etal-2020-evaluation}
Sergey Edunov, Myle Ott, Marc{'}Aurelio Ranzato, and Michael Auli. 2020.
\newblock \href {https://doi.org/10.18653/v1/2020.acl-main.253} {{On The
  Evaluation of Machine Translation Systems Trained With Back-Translation}}.
\newblock In \emph{Proceedings of the 58th Annual Meeting of the Association
  for Computational Linguistics}, pages 2836--2846, Online. Association for
  Computational Linguistics.

\bibitem[{Fadaee and Monz(2018)}]{fadaee-monz-2018-back}
Marzieh Fadaee and Christof Monz. 2018.
\newblock \href {https://doi.org/10.18653/v1/D18-1040} {{Back-Translation
  Sampling by Targeting Difficult Words in Neural Machine Translation}}.
\newblock In \emph{Proceedings of the 2018 Conference on Empirical Methods in
  Natural Language Processing}, pages 436--446, Brussels, Belgium. Association
  for Computational Linguistics.

\bibitem[{Felice et~al.(2016)Felice, Bryant, and
  Briscoe}]{felice-etal-2016-automatic}
Mariano Felice, Christopher Bryant, and Ted Briscoe. 2016.
\newblock \href {https://www.aclweb.org/anthology/C16-1079} {{Automatic
  Extraction of Learner Errors in {ESL} Sentences Using Linguistically Enhanced
  Alignments}}.
\newblock In \emph{Proceedings of the 26th International Conference on
  Computational Linguistics: Technical Papers}, pages 825--835, Osaka, Japan.
  The COLING 2016 Organizing Committee.

\bibitem[{Ge et~al.(2018{\natexlab{a}})Ge, Wei, and
  Zhou}]{ge-etal-2018-fluency}
Tao Ge, Furu Wei, and Ming Zhou. 2018{\natexlab{a}}.
\newblock \href {https://doi.org/10.18653/v1/P18-1097} {{Fluency Boost Learning
  and Inference for Neural Grammatical Error Correction}}.
\newblock In \emph{Proceedings of the 56th Annual Meeting of the Association
  for Computational Linguistics}, pages 1055--1065, Melbourne, Australia.
  Association for Computational Linguistics.

\bibitem[{Ge et~al.(2018{\natexlab{b}})Ge, Wei, and
  Zhou}]{ge-etal-2018-reaching}
Tao Ge, Furu Wei, and Ming Zhou. 2018{\natexlab{b}}.
\newblock \href {https://arxiv.org/abs/1807.01270} {{Reaching Human-level
  Performance in Automatic Grammatical Error Correction: An Empirical Study}}.
\newblock \emph{arXiv preprint arXiv:1807.01270v5 [cs.CL]}.

\bibitem[{Gehring et~al.(2017)Gehring, Auli, Grangier, Yarats, and
  Dauphin}]{gehring-etal-2017-convolutional}
Jonas Gehring, Michael Auli, David Grangier, Denis Yarats, and Yann~N. Dauphin.
  2017.
\newblock \href {http://proceedings.mlr.press/v70/gehring17a.html}
  {{Convolutional Sequence to Sequence Learning}}.
\newblock In \emph{Proceedings of the 34th International Conference on Machine
  Learning}, pages 1243--1252, Sydney, Australia. PMLR.

\bibitem[{Gra{\c{c}}a et~al.(2019)Gra{\c{c}}a, Kim, Schamper, Khadivi, and
  Ney}]{graca-etal-2019-generalizing}
Miguel Gra{\c{c}}a, Yunsu Kim, Julian Schamper, Shahram Khadivi, and Hermann
  Ney. 2019.
\newblock \href {https://doi.org/10.18653/v1/W19-5205} {{Generalizing
  Back-Translation in Neural Machine Translation}}.
\newblock In \emph{Proceedings of the Fourth Conference on Machine
  Translation}, pages 45--52, Florence, Italy. Association for Computational
  Linguistics.

\bibitem[{Granger(1998)}]{granger-1998-computerized}
Sylviane Granger. 1998.
\newblock \href {https://doi.org/https://doi.org/10.4324/9781315841342} {{The
  computerized learner corpus: a versatile new source of data for {SLA}
  research}}.
\newblock In \emph{Learner English on Computer}, pages 3--18. Addison Wesley
  Longman, London and New York.

\bibitem[{Grundkiewicz and
  Junczys-Dowmunt(2018)}]{grundkiewicz-junczys-dowmunt-2018-near}
Roman Grundkiewicz and Marcin Junczys-Dowmunt. 2018.
\newblock \href {https://doi.org/10.18653/v1/N18-2046} {{Near Human-Level
  Performance in Grammatical Error Correction with Hybrid Machine
  Translation}}.
\newblock In \emph{Proceedings of the 2018 Conference of the North {A}merican
  Chapter of the Association for Computational Linguistics: Human Language
  Technologies}, pages 284--290, New Orleans, Louisiana. Association for
  Computational Linguistics.

\bibitem[{Grundkiewicz and
  Junczys-Dowmunt(2019)}]{grundkiewicz-junczys-dowmunt-2019-minimally}
Roman Grundkiewicz and Marcin Junczys-Dowmunt. 2019.
\newblock \href {https://doi.org/10.18653/v1/D19-5546} {{Minimally-Augmented
  Grammatical Error Correction}}.
\newblock In \emph{Proceedings of the 5th Workshop on Noisy User-generated
  Text}, pages 357--363, Hong Kong, China. Association for Computational
  Linguistics.

\bibitem[{Grundkiewicz et~al.(2019)Grundkiewicz, Junczys-Dowmunt, and
  Heafield}]{grundkiewicz-etal-2019-neural}
Roman Grundkiewicz, Marcin Junczys-Dowmunt, and Kenneth Heafield. 2019.
\newblock \href {https://doi.org/10.18653/v1/W19-4427} {{Neural Grammatical
  Error Correction Systems with Unsupervised Pre-training on Synthetic Data}}.
\newblock In \emph{Proceedings of the Fourteenth Workshop on Innovative Use of
  NLP for Building Educational Applications}, pages 252--263, Florence, Italy.
  Association for Computational Linguistics.

\bibitem[{Heilman et~al.(2014)Heilman, Cahill, Madnani, Lopez, Mulholland, and
  Tetreault}]{heilman-etal-2014-predicting}
Michael Heilman, Aoife Cahill, Nitin Madnani, Melissa Lopez, Matthew
  Mulholland, and Joel Tetreault. 2014.
\newblock \href {https://doi.org/10.3115/v1/P14-2029} {{Predicting
  Grammaticality on an Ordinal Scale}}.
\newblock In \emph{Proceedings of the 52nd Annual Meeting of the Association
  for Computational Linguistics}, pages 174--180, Baltimore, Maryland.
  Association for Computational Linguistics.

\bibitem[{Hotate et~al.(2020)Hotate, Kaneko, and
  Komachi}]{hotate-etal-2020-generating}
Kengo Hotate, Masahiro Kaneko, and Mamoru Komachi. 2020.
\newblock \href {https://doi.org/10.18653/v1/2020.coling-main.193} {{Generating
  Diverse Corrections with Local Beam Search for Grammatical Error
  Correction}}.
\newblock In \emph{Proceedings of the 28th International Conference on
  Computational Linguistics}, pages 2132--2137, Barcelona, Spain (Online).
  International Committee on Computational Linguistics.

\bibitem[{Htut and Tetreault(2019)}]{htut-tetreault-2019-unbearable}
Phu~Mon Htut and Joel Tetreault. 2019.
\newblock \href {https://doi.org/10.18653/v1/W19-4449} {{The Unbearable Weight
  of Generating Artificial Errors for Grammatical Error Correction}}.
\newblock In \emph{Proceedings of the Fourteenth Workshop on Innovative Use of
  NLP for Building Educational Applications}, pages 478--483, Florence, Italy.
  Association for Computational Linguistics.

\bibitem[{Ji et~al.(2017)Ji, Wang, Toutanova, Gong, Truong, and
  Gao}]{ji-etal-2017-nested}
Jianshu Ji, Qinlong Wang, Kristina Toutanova, Yongen Gong, Steven Truong, and
  Jianfeng Gao. 2017.
\newblock \href {https://doi.org/10.18653/v1/P17-1070} {{A Nested Attention
  Neural Hybrid Model for Grammatical Error Correction}}.
\newblock In \emph{Proceedings of the 55th Annual Meeting of the Association
  for Computational Linguistics}, pages 753--762, Vancouver, Canada.
  Association for Computational Linguistics.

\bibitem[{Junczys-Dowmunt et~al.(2018)Junczys-Dowmunt, Grundkiewicz, Guha, and
  Heafield}]{junczys-dowmunt-etal-2018-approaching}
Marcin Junczys-Dowmunt, Roman Grundkiewicz, Shubha Guha, and Kenneth Heafield.
  2018.
\newblock \href {https://doi.org/10.18653/v1/N18-1055} {{Approaching Neural
  Grammatical Error Correction as a Low-Resource Machine Translation Task}}.
\newblock In \emph{Proceedings of the 2018 Conference of the North {A}merican
  Chapter of the Association for Computational Linguistics: Human Language
  Technologies}, pages 595--606, New Orleans, Louisiana. Association for
  Computational Linguistics.

\bibitem[{Kaneko et~al.(2020)Kaneko, Mita, Kiyono, Suzuki, and
  Inui}]{kaneko-etal-2020-encoder}
Masahiro Kaneko, Masato Mita, Shun Kiyono, Jun Suzuki, and Kentaro Inui. 2020.
\newblock \href {https://doi.org/10.18653/v1/2020.acl-main.391}
  {{Encoder-Decoder Models Can Benefit from Pre-trained Masked Language Models
  in Grammatical Error Correction}}.
\newblock In \emph{Proceedings of the 58th Annual Meeting of the Association
  for Computational Linguistics}, pages 4248--4254, Online. Association for
  Computational Linguistics.

\bibitem[{Kantor et~al.(2019)Kantor, Katz, Choshen, Cohen-Karlik, Liberman,
  Toledo, Menczel, and Slonim}]{kantor-etal-2019-learning}
Yoav Kantor, Yoav Katz, Leshem Choshen, Edo Cohen-Karlik, Naftali Liberman,
  Assaf Toledo, Amir Menczel, and Noam Slonim. 2019.
\newblock \href {https://doi.org/10.18653/v1/W19-4414} {{Learning to combine
  Grammatical Error Corrections}}.
\newblock In \emph{Proceedings of the Fourteenth Workshop on Innovative Use of
  NLP for Building Educational Applications}, pages 139--148, Florence, Italy.
  Association for Computational Linguistics.

\bibitem[{Kasewa et~al.(2018)Kasewa, Stenetorp, and
  Riedel}]{kasewa-etal-2018-wronging}
Sudhanshu Kasewa, Pontus Stenetorp, and Sebastian Riedel. 2018.
\newblock \href {https://doi.org/10.18653/v1/D18-1541} {{Wronging a Right:
  Generating Better Errors to Improve Grammatical Error Detection}}.
\newblock In \emph{Proceedings of the 2018 Conference on Empirical Methods in
  Natural Language Processing}, pages 4977--4983, Brussels, Belgium.
  Association for Computational Linguistics.

\bibitem[{Kingma and Ba(2015)}]{kingma-ba-2015-adam}
Diederik~P. Kingma and Jimmy Ba. 2015.
\newblock \href {http://arxiv.org/abs/1412.6980} {{Adam: {A} Method for
  Stochastic Optimization}}.
\newblock In \emph{Proceedings of the 3rd International Conference on Learning
  Representations}, San Diego, California.

\bibitem[{Kiyono et~al.(2019)Kiyono, Suzuki, Mita, Mizumoto, and
  Inui}]{kiyono-etal-2019-empirical}
Shun Kiyono, Jun Suzuki, Masato Mita, Tomoya Mizumoto, and Kentaro Inui. 2019.
\newblock \href {https://doi.org/10.18653/v1/D19-1119} {{An Empirical Study of
  Incorporating Pseudo Data into Grammatical Error Correction}}.
\newblock In \emph{Proceedings of the 2019 Conference on Empirical Methods in
  Natural Language Processing and the 9th International Joint Conference on
  Natural Language Processing}, pages 1236--1242, Hong Kong, China. Association
  for Computational Linguistics.

\bibitem[{Kiyono et~al.(2020)Kiyono, Suzuki, Mizumoto, and
  Inui}]{kiyono-etal-2020-massive}
Shun Kiyono, Jun Suzuki, Tomoya Mizumoto, and Kentaro Inui. 2020.
\newblock \href {https://doi.org/10.1109/TASLP.2020.3007753} {{Massive
  Exploration of Pseudo Data for Grammatical Error Correction}}.
\newblock \emph{IEEE/ACM Transactions on Audio, Speech, and Language
  Processing}, 28:2134--2145.

\bibitem[{Koehn and Knowles(2017)}]{koehn-knowles-2017-six}
Philipp Koehn and Rebecca Knowles. 2017.
\newblock \href {https://doi.org/10.18653/v1/W17-3204} {{Six Challenges for
  Neural Machine Translation}}.
\newblock In \emph{Proceedings of the First Workshop on Neural Machine
  Translation}, pages 28--39, Vancouver, Canada. Association for Computational
  Linguistics.

\bibitem[{Lichtarge et~al.(2019)Lichtarge, Alberti, Kumar, Shazeer, Parmar, and
  Tong}]{lichtarge-etal-2019-corpora}
Jared Lichtarge, Chris Alberti, Shankar Kumar, Noam Shazeer, Niki Parmar, and
  Simon Tong. 2019.
\newblock \href {https://doi.org/10.18653/v1/N19-1333} {{Corpora Generation for
  Grammatical Error Correction}}.
\newblock In \emph{Proceedings of the 2019 Conference of the North {A}merican
  Chapter of the Association for Computational Linguistics: Human Language
  Technologies}, pages 3291--3301, Minneapolis, Minnesota. Association for
  Computational Linguistics.

\bibitem[{Luong et~al.(2015)Luong, Pham, and
  Manning}]{luong-etal-2015-effective}
Thang Luong, Hieu Pham, and Christopher~D. Manning. 2015.
\newblock \href {https://doi.org/10.18653/v1/D15-1166} {{Effective Approaches
  to Attention-based Neural Machine Translation}}.
\newblock In \emph{Proceedings of the 2015 Conference on Empirical Methods in
  Natural Language Processing}, pages 1412--1421, Lisbon, Portugal. Association
  for Computational Linguistics.

\bibitem[{Mizumoto et~al.(2011)Mizumoto, Komachi, Nagata, and
  Matsumoto}]{mizumoto-etal-2011-mining}
Tomoya Mizumoto, Mamoru Komachi, Masaaki Nagata, and Yuji Matsumoto. 2011.
\newblock \href {https://www.aclweb.org/anthology/I11-1017} {{Mining Revision
  Log of Language Learning {SNS} for Automated {J}apanese Error Correction of
  Second Language Learners}}.
\newblock In \emph{Proceedings of 5th International Joint Conference on Natural
  Language Processing}, pages 147--155, Chiang Mai, Thailand. Asian Federation
  of Natural Language Processing.

\bibitem[{Napoles et~al.(2015)Napoles, Sakaguchi, Post, and
  Tetreault}]{napoles-etal-2015-ground}
Courtney Napoles, Keisuke Sakaguchi, Matt Post, and Joel Tetreault. 2015.
\newblock \href {https://doi.org/10.3115/v1/P15-2097} {{Ground Truth for
  Grammatical Error Correction Metrics}}.
\newblock In \emph{Proceedings of the 53rd Annual Meeting of the Association
  for Computational Linguistics and the 7th International Joint Conference on
  Natural Language Processing}, pages 588--593, Beijing, China. Association for
  Computational Linguistics.

\bibitem[{Napoles et~al.(2016)Napoles, Sakaguchi, Post, and
  Tetreault}]{napoles-etal-2016-gleu}
Courtney Napoles, Keisuke Sakaguchi, Matt Post, and Joel Tetreault. 2016.
\newblock \href {https://arxiv.org/abs/1605.02592} {{GLEU Without Tuning}}.
\newblock \emph{arXiv preprint arXiv:1605.02592v1 [cs.CL]}.

\bibitem[{Napoles et~al.(2017)Napoles, Sakaguchi, and
  Tetreault}]{napoles-etal-2017-jfleg}
Courtney Napoles, Keisuke Sakaguchi, and Joel Tetreault. 2017.
\newblock \href {https://www.aclweb.org/anthology/E17-2037} {{{JFLEG}: A
  Fluency Corpus and Benchmark for Grammatical Error Correction}}.
\newblock In \emph{Proceedings of the 15th Conference of the {E}uropean Chapter
  of the Association for Computational Linguistics}, pages 229--234, Valencia,
  Spain. Association for Computational Linguistics.

\bibitem[{Ng et~al.(2014)Ng, Wu, Briscoe, Hadiwinoto, Susanto, and
  Bryant}]{ng-etal-2014-conll}
Hwee~Tou Ng, Siew~Mei Wu, Ted Briscoe, Christian Hadiwinoto, Raymond~Hendy
  Susanto, and Christopher Bryant. 2014.
\newblock \href {https://doi.org/10.3115/v1/W14-1701} {{The {C}o{NLL}-2014
  Shared Task on Grammatical Error Correction}}.
\newblock In \emph{Proceedings of the Eighteenth Conference on Computational
  Natural Language Learning: Shared Task}, pages 1--14, Baltimore, Maryland.
  Association for Computational Linguistics.

\bibitem[{Ott et~al.(2019)Ott, Edunov, Baevski, Fan, Gross, Ng, Grangier, and
  Auli}]{ott-etal-2019-fairseq}
Myle Ott, Sergey Edunov, Alexei Baevski, Angela Fan, Sam Gross, Nathan Ng,
  David Grangier, and Michael Auli. 2019.
\newblock \href {https://doi.org/10.18653/v1/N19-4009} {{fairseq: A Fast,
  Extensible Toolkit for Sequence Modeling}}.
\newblock In \emph{Proceedings of the 2019 Conference of the North {A}merican
  Chapter of the Association for Computational Linguistics (Demonstrations)},
  pages 48--53, Minneapolis, Minnesota. Association for Computational
  Linguistics.

\bibitem[{Poncelas et~al.(2018)Poncelas, Shterionov, Way, {De Buy Wenniger},
  and Passban}]{poncelas-etal-2018-investigating}
Alberto Poncelas, Dimitar Shterionov, Andy Way, {Gideon Maillette} {De Buy
  Wenniger}, and Peyman Passban. 2018.
\newblock \href {http://rua.ua.es/dspace/handle/10045/76085?locale=en}
  {{Investigating Backtranslation in Neural Machine Translation}}.
\newblock In \emph{Proceedings of the 21st Annual Conference of the European
  Association for Machine Translation}, pages 249--258, Alacant, Spain.
  European Association for Machine Translation.

\bibitem[{Qiu et~al.(2019)Qiu, Chen, Liu, Parvathala, Patil, and
  Park}]{qiu-etal-2019-improving}
Mengyang Qiu, Xuejiao Chen, Maggie Liu, Krishna Parvathala, Apurva Patil, and
  Jungyeul Park. 2019.
\newblock \href {https://doi.org/10.18653/v1/W19-4425} {{Improving Precision of
  Grammatical Error Correction with a Cheat Sheet}}.
\newblock In \emph{Proceedings of the Fourteenth Workshop on Innovative Use of
  NLP for Building Educational Applications}, pages 240--245, Florence, Italy.
  Association for Computational Linguistics.

\bibitem[{Rei et~al.(2017)Rei, Felice, Yuan, and
  Briscoe}]{rei-etal-2017-artificial}
Marek Rei, Mariano Felice, Zheng Yuan, and Ted Briscoe. 2017.
\newblock \href {https://doi.org/10.18653/v1/W17-5032} {{Artificial Error
  Generation with Machine Translation and Syntactic Patterns}}.
\newblock In \emph{Proceedings of the 12th Workshop on Innovative Use of {NLP}
  for Building Educational Applications}, pages 287--292, Copenhagen, Denmark.
  Association for Computational Linguistics.

\bibitem[{Sennrich et~al.(2016{\natexlab{a}})Sennrich, Haddow, and
  Birch}]{sennrich-etal-2016-improving}
Rico Sennrich, Barry Haddow, and Alexandra Birch. 2016{\natexlab{a}}.
\newblock \href {https://doi.org/10.18653/v1/P16-1009} {{Improving Neural
  Machine Translation Models with Monolingual Data}}.
\newblock In \emph{Proceedings of the 54th Annual Meeting of the Association
  for Computational Linguistics}, pages 86--96, Berlin, Germany. Association
  for Computational Linguistics.

\bibitem[{Sennrich et~al.(2016{\natexlab{b}})Sennrich, Haddow, and
  Birch}]{sennrich-etal-2016-neural}
Rico Sennrich, Barry Haddow, and Alexandra Birch. 2016{\natexlab{b}}.
\newblock \href {https://doi.org/10.18653/v1/P16-1162} {{Neural Machine
  Translation of Rare Words with Subword Units}}.
\newblock In \emph{Proceedings of the 54th Annual Meeting of the Association
  for Computational Linguistics}, pages 1715--1725, Berlin, Germany.
  Association for Computational Linguistics.

\bibitem[{Shazeer and Stern(2018)}]{shazeer-stern-2018-adafactor}
Noam Shazeer and Mitchell Stern. 2018.
\newblock \href {http://proceedings.mlr.press/v80/shazeer18a.html} {{Adafactor:
  Adaptive Learning Rates with Sublinear Memory Cost}}.
\newblock In \emph{Proceedings of the 35th International Conference on Machine
  Learning}, pages 4596--4604, Stockholm, Sweden. PMLR.

\bibitem[{Shen et~al.(2018)Shen, Lin, Huang, and
  Courville}]{shen-etal-2018-neural}
Yikang Shen, Zhouhan Lin, Chin{-}Wei Huang, and Aaron~C. Courville. 2018.
\newblock \href {https://openreview.net/forum?id=rkgOLb-0W} {{Neural Language
  Modeling by Jointly Learning Syntax and Lexicon}}.
\newblock In \emph{Proceedings of the 6th International Conference on Learning
  Representations}, Vancouver, Canada. OpenReview.net.

\bibitem[{Shen et~al.(2019)Shen, Tan, Sordoni, and
  Courville}]{shen-etal-2019-ordered}
Yikang Shen, Shawn Tan, Alessandro Sordoni, and Aaron~C. Courville. 2019.
\newblock \href {https://openreview.net/forum?id=B1l6qiR5F7} {{Ordered Neurons:
  Integrating Tree Structures into Recurrent Neural Networks}}.
\newblock In \emph{Proceedings of the 7th International Conference on Learning
  Representations}, New Orleans, Louisiana. OpenReview.net.

\bibitem[{Soto et~al.(2020)Soto, Shterionov, Poncelas, and
  Way}]{soto-etal-2020-selecting}
Xabier Soto, Dimitar Shterionov, Alberto Poncelas, and Andy Way. 2020.
\newblock \href {https://doi.org/10.18653/v1/2020.acl-main.359} {{Selecting
  Backtranslated Data from Multiple Sources for Improved Neural Machine
  Translation}}.
\newblock In \emph{Proceedings of the 58th Annual Meeting of the Association
  for Computational Linguistics}, pages 3898--3908, Online. Association for
  Computational Linguistics.

\bibitem[{Srivastava et~al.(2014)Srivastava, Hinton, Krizhevsky, Sutskever, and
  Salakhutdinov}]{srivastava-etal-2014-dropout}
Nitish Srivastava, Geoffrey Hinton, Alex Krizhevsky, Ilya Sutskever, and Ruslan
  Salakhutdinov. 2014.
\newblock \href {http://jmlr.org/papers/v15/srivastava14a.html} {{Dropout: A
  Simple Way to Prevent Neural Networks from Overfitting}}.
\newblock \emph{Journal of Machine Learning Research}, 15(56):1929--1958.

\bibitem[{Sutskever et~al.(2014)Sutskever, Vinyals, and
  Le}]{sutskever-etal-2014-sequence}
Ilya Sutskever, Oriol Vinyals, and Quoc~V Le. 2014.
\newblock \href
  {https://proceedings.neurips.cc/paper/2014/file/a14ac55a4f27472c5d894ec1c3c743d2-Paper.pdf}
  {{Sequence to Sequence Learning with Neural Networks}}.
\newblock In \emph{Advances in Neural Information Processing Systems 27}, pages
  3104--3112, Montreal, Canada. Curran Associates, Inc.

\bibitem[{Szegedy et~al.(2016)Szegedy, Vanhoucke, Ioffe, Shlens, and
  Wojna}]{szegedy-etal-2016-rethinking}
Christian Szegedy, Vincent Vanhoucke, Sergey Ioffe, Jon Shlens, and Zbigniew
  Wojna. 2016.
\newblock \href {https://doi.org/10.1109/CVPR.2016.308} {{Rethinking the
  Inception Architecture for Computer Vision}}.
\newblock In \emph{Proceedings of the 2016 IEEE Conference on Computer Vision
  and Pattern Recognition}, pages 2818--2826, Las Vegas, Nevada. Institute of
  Electrical and Electronics Engineers.

\bibitem[{Tajiri et~al.(2012)Tajiri, Komachi, and
  Matsumoto}]{tajiri-etal-2012-tense}
Toshikazu Tajiri, Mamoru Komachi, and Yuji Matsumoto. 2012.
\newblock \href {https://www.aclweb.org/anthology/P12-2039} {{Tense and Aspect
  Error Correction for {ESL} Learners Using Global Context}}.
\newblock In \emph{Proceedings of the 50th Annual Meeting of the Association
  for Computational Linguistics}, pages 198--202, Jeju Island, Korea.
  Association for Computational Linguistics.

\bibitem[{Takahashi et~al.(2020)Takahashi, Katsumata, and
  Komachi}]{takahashi-etal-2020-grammatical}
Yujin Takahashi, Satoru Katsumata, and Mamoru Komachi. 2020.
\newblock \href {https://doi.org/10.18653/v1/2020.acl-srw.5} {{Grammatical
  Error Correction Using Pseudo Learner Corpus Considering Learner{'}s Error
  Tendency}}.
\newblock In \emph{Proceedings of the 58th Annual Meeting of the Association
  for Computational Linguistics: Student Research Workshop}, pages 27--32,
  Online. Association for Computational Linguistics.

\bibitem[{Vaswani et~al.(2017)Vaswani, Shazeer, Parmar, Uszkoreit, Jones,
  Gomez, Kaiser, and Polosukhin}]{vaswani-etal-2017-attention}
Ashish Vaswani, Noam Shazeer, Niki Parmar, Jakob Uszkoreit, Llion Jones,
  Aidan~N Gomez, {\L}ukasz Kaiser, and Illia Polosukhin. 2017.
\newblock \href
  {https://proceedings.neurips.cc/paper/2017/file/3f5ee243547dee91fbd053c1c4a845aa-Paper.pdf}
  {{Attention is All you Need}}.
\newblock In \emph{Advances in Neural Information Processing Systems 30}, pages
  5998--6008, Long Beach, California. Curran Associates, Inc.

\bibitem[{Wan et~al.(2020)Wan, Wan, and Wang}]{wan-etal-2020-improving}
Zhaohong Wan, Xiaojun Wan, and Wenguang Wang. 2020.
\newblock \href {https://doi.org/10.18653/v1/2020.coling-main.200} {{Improving
  Grammatical Error Correction with Data Augmentation by Editing Latent
  Representation}}.
\newblock In \emph{Proceedings of the 28th International Conference on
  Computational Linguistics}, pages 2202--2212, Barcelona, Spain (Online).
  International Committee on Computational Linguistics.

\bibitem[{Wang et~al.(2020)Wang, Yang, Chen, Du, and
  Yang}]{wang-etal-2020-controllable}
Chencheng Wang, Liner Yang, Yun Chen, Yongping Du, and Erhong Yang. 2020.
\newblock \href {https://arxiv.org/abs/1909.13302} {{Controllable Data
  Synthesis Method for Grammatical Error Correction}}.
\newblock \emph{arXiv preprint arXiv:1909.13302v3 [cs.CL]}.

\bibitem[{Wang and Zheng(2020)}]{wang-zheng-2020-improving}
Lihao Wang and Xiaoqing Zheng. 2020.
\newblock \href {https://doi.org/10.18653/v1/2020.emnlp-main.228} {{Improving
  Grammatical Error Correction Models with Purpose-Built Adversarial
  Examples}}.
\newblock In \emph{Proceedings of the 2020 Conference on Empirical Methods in
  Natural Language Processing}, pages 2858--2869, Online. Association for
  Computational Linguistics.

\bibitem[{White and Rozovskaya(2020)}]{white-rozovskaya-2020-comparative}
Max White and Alla Rozovskaya. 2020.
\newblock \href {https://doi.org/10.18653/v1/2020.bea-1.21} {{A Comparative
  Study of Synthetic Data Generation Methods for Grammatical Error
  Correction}}.
\newblock In \emph{Proceedings of the Fifteenth Workshop on Innovative Use of
  NLP for Building Educational Applications}, pages 198--208, Seattle,
  Washington (Online). Association for Computational Linguistics.

\bibitem[{Xie et~al.(2018)Xie, Genthial, Xie, Ng, and
  Jurafsky}]{xie-etal-2018-noising}
Ziang Xie, Guillaume Genthial, Stanley Xie, Andrew Ng, and Dan Jurafsky. 2018.
\newblock \href {https://doi.org/10.18653/v1/N18-1057} {{Noising and Denoising
  Natural Language: Diverse Backtranslation for Grammar Correction}}.
\newblock In \emph{Proceedings of the 2018 Conference of the North {A}merican
  Chapter of the Association for Computational Linguistics: Human Language
  Technologies}, pages 619--628, New Orleans, Louisiana. Association for
  Computational Linguistics.

\bibitem[{Xu et~al.(2019)Xu, Zhang, Chen, and Qin}]{xu-etal-2019-erroneous}
Shuyao Xu, Jiehao Zhang, Jin Chen, and Long Qin. 2019.
\newblock \href {https://doi.org/10.18653/v1/W19-4415} {{Erroneous data
  generation for Grammatical Error Correction}}.
\newblock In \emph{Proceedings of the Fourteenth Workshop on Innovative Use of
  NLP for Building Educational Applications}, pages 149--158, Florence, Italy.
  Association for Computational Linguistics.

\bibitem[{Yannakoudakis et~al.(2011)Yannakoudakis, Briscoe, and
  Medlock}]{yannakoudakis-etal-2011-new}
Helen Yannakoudakis, Ted Briscoe, and Ben Medlock. 2011.
\newblock \href {https://www.aclweb.org/anthology/P11-1019} {{A New Dataset and
  Method for Automatically Grading {ESOL} Texts}}.
\newblock In \emph{Proceedings of the 49th Annual Meeting of the Association
  for Computational Linguistics: Human Language Technologies}, pages 180--189,
  Portland, Oregon. Association for Computational Linguistics.

\bibitem[{Yannakoudakis et~al.(2018)Yannakoudakis, Øistein E~Andersen,
  Geranpayeh, Briscoe, and Nicholls}]{yannakoudakis-etal-2018-developing}
Helen Yannakoudakis, Øistein E~Andersen, Ardeshir Geranpayeh, Ted Briscoe, and
  Diane Nicholls. 2018.
\newblock \href {https://doi.org/10.1080/08957347.2018.1464447} {{Developing an
  automated writing placement system for ESL learners}}.
\newblock \emph{Applied Measurement in Education}, 31(3):251--267.

\bibitem[{Yuan and Briscoe(2016)}]{yuan-briscoe-2016-grammatical}
Zheng Yuan and Ted Briscoe. 2016.
\newblock \href {https://doi.org/10.18653/v1/N16-1042} {{Grammatical error
  correction using neural machine translation}}.
\newblock In \emph{Proceedings of the 2016 Conference of the North {A}merican
  Chapter of the Association for Computational Linguistics: Human Language
  Technologies}, pages 380--386, San Diego, California. Association for
  Computational Linguistics.

\bibitem[{Zhang et~al.(2019)Zhang, Ge, Wei, Zhou, and
  Sun}]{zhang-etal-2019-sequencetosequence}
Yi~Zhang, Tao Ge, Furu Wei, Ming Zhou, and Xu~Sun. 2019.
\newblock \href {https://arxiv.org/abs/1909.06002} {{Sequence-to-sequence
  Pre-training with Data Augmentation for Sentence Rewriting}}.
\newblock \emph{arXiv preprint arXiv:1909.06002v2 [cs.CL]}.

\bibitem[{Zhao et~al.(2019)Zhao, Wang, Shen, Jia, and
  Liu}]{zhao-etal-2019-improving}
Wei Zhao, Liang Wang, Kewei Shen, Ruoyu Jia, and Jingming Liu. 2019.
\newblock \href {https://doi.org/10.18653/v1/N19-1014} {{Improving Grammatical
  Error Correction via Pre-Training a Copy-Augmented Architecture with
  Unlabeled Data}}.
\newblock In \emph{Proceedings of the 2019 Conference of the North {A}merican
  Chapter of the Association for Computational Linguistics: Human Language
  Technologies}, pages 156--165, Minneapolis, Minnesota. Association for
  Computational Linguistics.

\bibitem[{Zhou et~al.(2020{\natexlab{a}})Zhou, Ge, Mu, Xu, Wei, and
  Zhou}]{zhou-etal-2020-improving-grammatical}
Wangchunshu Zhou, Tao Ge, Chang Mu, Ke~Xu, Furu Wei, and Ming Zhou.
  2020{\natexlab{a}}.
\newblock \href {https://doi.org/10.18653/v1/2020.findings-emnlp.30}
  {{Improving Grammatical Error Correction with Machine Translation Pairs}}.
\newblock In \emph{Findings of the Association for Computational Linguistics:
  EMNLP 2020}, pages 318--328, Online. Association for Computational
  Linguistics.

\bibitem[{Zhou et~al.(2020{\natexlab{b}})Zhou, Ge, and
  Xu}]{zhou-etal-2020-pseudo}
Wangchunshu Zhou, Tao Ge, and Ke~Xu. 2020{\natexlab{b}}.
\newblock \href {https://doi.org/10.18653/v1/2020.findings-emnlp.136}
  {{Pseudo-Bidirectional Decoding for Local Sequence Transduction}}.
\newblock In \emph{Findings of the Association for Computational Linguistics:
  EMNLP 2020}, pages 1506--1511, Online. Association for Computational
  Linguistics.

\end{thebibliography}
\bibliographystyle{acl_natbib}

\end{document}